\documentclass[11pt]{article}

\usepackage{acl}

\usepackage{times}
\usepackage{latexsym}
\usepackage[T1]{fontenc}
\usepackage[utf8]{inputenc}
\usepackage{microtype}
\usepackage{inconsolata}
\usepackage{graphicx}
\usepackage{booktabs}
\usepackage{amsmath}
\usepackage{xcolor}
\usepackage{tikz}
\usetikzlibrary{shapes,arrows.meta,positioning,fit,backgrounds}

\usepackage{enumitem}
\setlist{nosep,leftmargin=*}
\usepackage[breakable,skins]{tcolorbox}
\usepackage{float}

\definecolor{clrSFT}{HTML}{E8A087}
\definecolor{clrCls}{HTML}{88C4C8}
\definecolor{clrEns}{HTML}{D1BC8A}

\title{N\"urnberg NLP at PsyDefDetect: Multi-Axis Voter Ensembles\\for Psychological Defence Mechanism Classification}

\author{Philipp Steigerwald, Eric Rudolph \and Jens Albrecht \\
  Technische Hochschule N\"urnberg Georg Simon Ohm \\
  \texttt{\{philipp.steigerwald,eric.rudolph,jens.albrecht\}@th-nuernberg.de}}

\begin{document}
\maketitle

\begin{abstract}
Detecting levels of psychological defence mechanisms in supportive conversations is inherently ambiguous.
In the PsyDefDetect shared task at BioNLP 2026 the eight positive defence categories share surface language and differ only in pragmatic function and trained raters reach only moderate inter-annotator agreement.
On such a task the decisive lever is not a stronger single model but error independence, since any single representation will waver on the overlapping defence boundaries.
We translate this insight into a 9-voter ensemble spanning three orthogonal axes: class granularity (all nine classes for the gatekeeper, only the eight defence classes for the specialists), training method (generative and discriminative) and base model.
The system reaches $F1_{test}{=}.420$ on the hidden test set, placing first among 21 registered teams.
\end{abstract}

\section{Introduction}
\label{sec:intro}

The PsyDefDetect shared task \citep{na-etal-2026-psydefdetect} asks a model to classify each seeker utterance in an emotional-support conversation by its level of psychological defence.
The PSYDEFCONV corpus \citep{na-etal-2026-psydefconv} pairs ESConv \citep{liu2021esconv}, a corpus of crowdsourced support dialogues, with the Defense Mechanism Rating Scale \citep[DMRS;][]{perry1990defense}---a clinical taxonomy of eight hierarchical defence levels plus a ``No Defence'' category---and is evaluated on macro-F1 over classes 1--8.
The task is clinically motivated \citep{perry2014anomalies} but difficult: trained raters reach only a moderate Cohen's $\kappa{=}.639$ and the corpus is heavily imbalanced (C7 covers $52\%$, the three rarest classes together $7.4\%$).
Several defence categories share surface-level language and differ only in pragmatic function, so the semantic boundaries are inherently fuzzy.

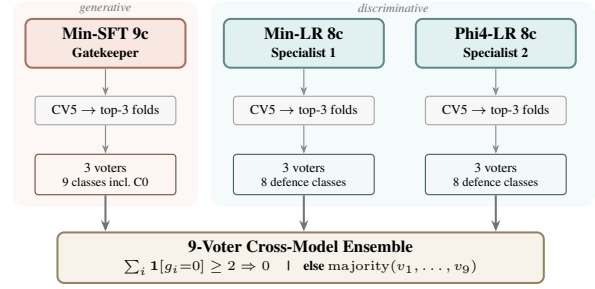
\begin{figure}[t]
\centering
\resizebox{\columnwidth}{!}{%
\begin{tikzpicture}[
    module/.style={rectangle, minimum width=2.5cm, minimum height=0.7cm,
        align=center, font=\fontsize{7.5}{9}\selectfont\bfseries,
        rounded corners=3pt, line width=0.9pt},
    detail/.style={rectangle, draw=black!35, fill=black!3,
        minimum width=2.2cm, minimum height=0.42cm,
        align=center, font=\fontsize{6}{7.5}\selectfont,
        rounded corners=1.5pt, line width=0.5pt},
    voterbox/.style={rectangle, minimum width=2.2cm, minimum height=0.42cm,
        align=center, font=\fontsize{6}{7.5}\selectfont,
        rounded corners=1.5pt, line width=0.5pt},
    rolelabel/.style={font=\fontsize{5.5}{7}\selectfont\itshape, text=black!55},
    arrow/.style={-{Stealth[length=4pt]}, line width=0.5pt, black!45},
    pipearrow/.style={-{Stealth[length=5pt, width=4pt]}, line width=0.9pt, black!55},
    groupbox/.style={rectangle, rounded corners=5pt, line width=0.6pt, dashed},
]
\begin{scope}[on background layer]
  \fill[clrSFT!8, rounded corners=5pt] (-4.05, 0.65) rectangle (-1.15, -2.55);
  \fill[clrCls!8, rounded corners=5pt] (-0.95, 0.65) rectangle (5.05, -2.55);
\end{scope}
\node[rolelabel] at (-2.6, 0.55) {generative};
\node[rolelabel] at ( 1.9, 0.55) {discriminative};
\node[module, draw=clrSFT!70!black, fill=clrSFT!25] (sft) at (-2.6, 0)
    {Min-SFT 9c\\[-1pt]{\fontsize{6}{7.5}\selectfont Gatekeeper}};
\node[module, draw=clrCls!70!black, fill=clrCls!25] (lr1) at (0.5, 0)
    {Min-LR 8c\\[-1pt]{\fontsize{6}{7.5}\selectfont Specialist 1}};
\node[module, draw=clrCls!70!black, fill=clrCls!25] (lr2) at (3.5, 0)
    {Phi4-LR 8c\\[-1pt]{\fontsize{6}{7.5}\selectfont Specialist 2}};
\node[detail] (sft_cv) at (-2.6, -1.05) {CV5 $\to$ top-3 folds};
\node[detail] (lr1_cv) at (0.5, -1.05) {CV5 $\to$ top-3 folds};
\node[detail] (lr2_cv) at (3.5, -1.05) {CV5 $\to$ top-3 folds};
\node[voterbox, draw=clrSFT!50!black, fill=clrSFT!10] (sft_v) at (-2.6, -2.05)
    {3 voters\\[-1pt]{\fontsize{5.5}{7}\selectfont 9 classes incl.\ C0}};
\node[voterbox, draw=clrCls!50!black, fill=clrCls!10] (lr1_v) at (0.5, -2.05)
    {3 voters\\[-1pt]{\fontsize{5.5}{7}\selectfont 8 defence classes}};
\node[voterbox, draw=clrCls!50!black, fill=clrCls!10] (lr2_v) at (3.5, -2.05)
    {3 voters\\[-1pt]{\fontsize{5.5}{7}\selectfont 8 defence classes}};
\draw[arrow] (sft.south) -- (sft_cv.north);
\draw[arrow] (lr1.south) -- (lr1_cv.north);
\draw[arrow] (lr2.south) -- (lr2_cv.north);
\draw[arrow] (sft_cv.south) -- (sft_v.north);
\draw[arrow] (lr1_cv.south) -- (lr1_v.north);
\draw[arrow] (lr2_cv.south) -- (lr2_v.north);
\node[module, draw=clrEns!70!black, fill=clrEns!18,
      minimum width=7.6cm, minimum height=0.78cm] (ens) at (0.45, -3.35)
    {9-Voter Cross-Model Ensemble\\[-1pt]{\fontsize{6}{7.5}\selectfont
     $\sum_{i} \mathbf{1}[g_i{=}0] \geq 2 \Rightarrow 0$ \quad\textbar\quad else $\operatorname{majority}(v_1,\dots,v_9)$}};
\draw[pipearrow] (sft_v.south) -- (sft_v.south |- ens.north);
\draw[pipearrow] (lr1_v.south) -- (lr1_v.south |- ens.north);
\draw[pipearrow] (lr2_v.south) -- (lr2_v.south |- ens.north);
\end{tikzpicture}%
}
\caption{Architecture of our 9-voter cross-model ensemble.}
\label{fig:system}
\end{figure}

Given these fuzzy boundaries, voting across diverse voters was our starting point. Rather than chasing a stronger single model, we sought voters with uncorrelated errors that arbitrate the ambiguity.
We tested different training methods (generative and discriminative), several base LLMs and different class granularities. A geometric analysis of the QLoRA-adapted hidden states indicates that only the no-defence class is reliably separable, motivating a \textit{generalist-specialist} split (9-class generalist + 8-class specialists).
To counter the heavy imbalance, we additionally augmented the minority defence classes with GPT-5.2 synthetic dialogues.

All training uses 5-fold cross-validation (CV5) as both a voter pool and an internal performance estimate ($F1_{cv}$, mean macro-F1 over classes 1--8) since test labels were hidden.
Comparing voters across folds, methods and base models, each axis produces systematically different errors: where some voters get confused on the fuzzy defence boundaries, others succeed and majority voting sharpens those boundaries---the error independence \citep{dietterich2000ensemble} we sought, realised in a 9-voter cross-model cross-method ensemble (Figure~\ref{fig:system}) that reaches $F1_{test}{=}.420$ ($+33.4\%$ over the baseline).

Our contributions: (i)~the winning 9-voter cross-model cross-method ensemble; and (ii)~an embedding-level analysis quantifying the defence-class semantic overlap that drives task difficulty. Additionally, we release our class-imbalance synthetic dialogues for replication.

\section{System}
\label{sec:system}

We build our 9-voter ensemble (Figure~\ref{fig:system}) step by step, adding one voter voice at a time. Each step posed a design choice---which method, which base model, which folds to trust---which we settled by the CV5 signal, guided by the principle that a diversity of voices sharpens fuzzy class boundaries better than any single strong voice.

\subsection{Data Augmentation}

The PSYDEFCONV training set is heavily imbalanced (C7 covers $52\%$, the three rarest classes together $7.4\%$).
Our first step was to replicate the organisers' baseline on a dialog-stratified 80/20 split, and the resulting 1{,}520-sample training split was augmented with up to $\min(200{-}n_c,\; 3 \cdot n_c)$ GPT-5.2 synthetic dialogues per class---the first term targets 200, the second caps synthetic at $75\%$.
Classes~0 and~7 are excluded as already well-represented; this yields 738 synthetic dialogues. When we later moved to CV5 for the voter pool, we reused the same 738 synthetic dialogues unchanged across all five folds (per-class counts in Appendix Table~\ref{tab:augmentation}).
Validation and test sets (472 samples) remain original human-annotated data only.

\subsection{Voting}
\label{sec:voting}

On this augmented data, voting was our first step toward the voter diversity we kept extending throughout the system.
Given the dataset's fuzzy class boundaries, the decisive lever is error independence between voters rather than a stronger single model; where one voter wavers, another may be more confident on the same sample and majority voting sharpens the joint decision.
CV5 provides both a voter pool (five trained models per configuration) and an internal performance estimate ($F1_{cv}$, mean macro-F1 over classes~1--8) since test labels were hidden, with the majority across the five fold-models giving the ensemble prediction.
Our first ensemble ran the organisers' baseline approach (Ministral-8B with generative supervised fine-tuning on all 9 classes of the augmented data, Min-SFT\,9c) as 5-fold majority voting, lifting $F1_{test}$ from $.315$ to $.373$ (Table~\ref{tab:progression})---already a substantial gain with no architectural diversity yet present.

\subsection{Training Axis}
\label{sec:training_axis}

We tested three adaptation methods for fine-tuning a base LLM.
\textbf{SFT} (supervised fine-tuning) fine-tuned the LLM end-to-end with QLoRA on a generative objective, learning to emit the class digit as text.
\textbf{ClsHead} (classification head) attached a randomly initialised head to the base LLM and jointly fine-tuned both with QLoRA and focal loss.
\textbf{LR} (logistic regression) froze the ClsHead-adapted LLM, discarded the trained classification head and fitted a new linear head on the frozen last-token hidden states---architecturally identical to the discarded one but retrained as an L2-regularised logistic regression on frozen features rather than jointly with the backbone.
Because LR reused the ClsHead-adapted backbone (extraction in minutes, fit in seconds), it added essentially no compute and let us screen many base-model and class-mode combinations cheaply. All three shared the same input prompt (see appendix).

\subsection{Class Granularity Axis}
\label{sec:class_axis}

To strengthen the 5-voter baseline we faced two design questions, which classes are reliably distinguishable in the hidden-state space and which configurations to combine for uncorrelated errors.

\begin{figure}[t]
\centering
\includegraphics[width=\columnwidth]{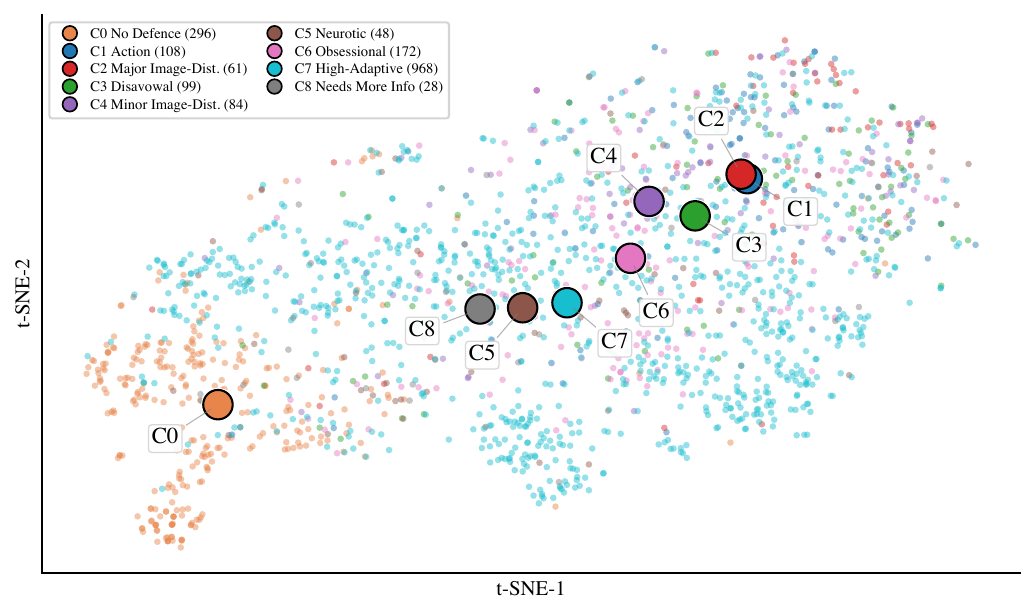}
\caption{Out-of-fold per-class t-SNE of SFT QLoRA-adapted 9-class Ministral-8B hidden states on the 1{,}864 original training utterances, with C0~No~Defence forming the only well-separated cluster.}
\label{fig:embeddings}
\end{figure}

A t-SNE of the SFT QLoRA-adapted 9-class Ministral-8B hidden states (Figure~\ref{fig:embeddings}) indicates that C0~No~Defence forms the most separable cluster, while the eight defence classes overlap substantially.
This motivates a class granularity split. The \emph{gatekeeper} keeps all nine classes, using C0 predictions for the no-defence override and C1--C8 predictions for the defence vote. The 8-class \emph{specialists} focus entirely on the overlapping defences.

For the 8-class specialist we tested seven base models with ClsHead and LR (Appendix Table~\ref{tab:cv5_models}).
LR matched or outperformed ClsHead in most cells and achieved the highest $F1_{cv}$ on the majority of base models, becoming the specialist default.
For the gatekeeper, although LR\,9c slightly outperforms SFT\,9c on $F1_{cv}$, we deliberately chose generative SFT, expecting two different methods to disagree on a different subset of samples than two LR branches would and trading a small per-voter $F1_{cv}$ loss for a larger gain in ensemble error independence.
A post-hoc ablation supports this. Pairing the Min-SFT\,9c gatekeeper with another SFT specialist (Min-SFT\,8c, top-3 folds, $t{=}2$) yields $F1_{test}{=}.373$, no improvement over the 5V baseline, while pairing with the discriminative Min-LR\,8c lifts it to $.391$ (Table~\ref{tab:progression}).

We paired the Min-SFT\,9c gatekeeper with the Min-LR\,8c specialist, keeping only the top-3 folds by $F1_{cv}$ per branch to drop each branch's most uncertain folds and save $40\%$ inference cost.
The resulting 6-voter ensemble fortuitously covers all five folds (Min-SFT\,9c $\{f0,f1,f4\}$, Min-LR\,8c $\{f0,f2,f3\}$).

With a dedicated gatekeeper, the simple majority vote extends to a two-stage rule.
The gatekeeper voters first decide whether the sample is C0 and the remaining defence classification is settled by majority across all voters.
Letting $g_1,\dots,g_G$ denote the $G$ gatekeeper predictions and $v_1,\dots,v_V$ all $V$ voter predictions, the ensemble decision is
\vspace{-2pt}
\begin{equation}
  \hat{y} =
  \begin{cases}
    0 \;\; \text{if} \; \textstyle\sum\nolimits_{i}^{G} \mathbf{1}[g_i{=}0] \geq (G{+}1)/2 \\[3pt]
    \operatorname*{argmax}_{c} \textstyle\sum\nolimits_{j}^{V} \mathbf{1}[v_j{=}c] \;\; \text{else}
  \end{cases}
  \label{eq:voting}
\end{equation}
\vspace{-2pt}
\vspace{2pt}
\noindent with ties broken in favour of class~7 (the majority class).
The gatekeeper voters participate in both branches of Equation~\ref{eq:voting}---triggering the C0-override when a majority of gatekeepers predicts C0, otherwise voting on defence classes alongside the LR specialists and adding method diversity since SFT and LR fail on different subsets of ambiguous samples.

\subsection{Model Axis}
\label{sec:model_axis}

To extend the diversity principle to the third (model) axis, we tested three additional 8-class LR variants---Phi-4-14B (Phi4-LR\,8c), Llama-3.1-8B (Llama-LR\,8c) and PsychoCounsel-Llama3-8B (PCounsel-LR\,8c, a counselling-domain Llama3-8B finetune)---and ranked them by per-fold Pearson correlation with Min-LR\,8c's $F1_{cv}$ profile (negative values indicate anti-aligned per-fold strengths, contributing independent voter signal).
We selected Phi4-LR\,8c on the most anti-aligned per-fold profile ($r{=}{-}.544$; Llama-LR\,8c $+.06$, PCounsel-LR\,8c $-.09$), completing the three-axis 9-voter ensemble.
With only $n{=}5$ folds, the gap between Phi4-LR ($r{=}{-}.544$) and Llama-LR ($r{=}{+}.06$) is indicative rather than statistically decisive; consistent with this, all three candidates land within $.006$ of each other on the test set (Table~\ref{tab:progression}).

\section{Results and Analysis}
\label{sec:results}

\begin{table}[t]
  \centering
  \footnotesize
  \setlength{\tabcolsep}{3pt}
  \begin{tabular*}{\columnwidth}{@{\extracolsep{\fill}}lc@{}}
    \toprule
    \textbf{System} & $\boldsymbol{F1_{test}}$ \\
    \midrule
    Baseline \citep{na-etal-2026-psydefconv} (Min-SFT\,9c, no-aug)  & .315 \\
    Min-SFT\,9c full-train, augmented (single model)         & .307 \\
    \midrule
    \multicolumn{2}{@{}l}{\textit{Voting baseline (no axes)}} \\
    \quad 5V Min-SFT\,9c (5 folds)                           & .373 \\
    \midrule
    \multicolumn{2}{@{}l}{\textit{Class + training axis}} \\
    \quad 6V Min-SFT\,9c + Min-LR\,8c                        & .391 \\
    \midrule
    \multicolumn{2}{@{}l}{\textit{Class + training + model axis}} \\
    \quad 6V Min-SFT\,9c + Phi4-LR\,8c                       & .391 \\
    \quad 6V Min-SFT\,9c + Llama-LR\,8c                      & .392 \\
    \quad 9V Min-SFT\,9c + Min-LR\,8c + PCounsel-LR\,8c      & .414 \\
    \quad 9V Min-SFT\,9c + Min-LR\,8c + Llama-LR\,8c         & .417 \\
    \quad \textbf{9V Min-SFT\,9c + Min-LR\,8c + Phi4-LR\,8c} & \textbf{.420} \\
    \bottomrule
  \end{tabular*}
  \caption{Hidden-test-set scores ($F1_{test}$, classes 1--8) for our submitted systems (all trained on augmented data), grouped by which diversity axes are active.}
  \label{tab:progression}
\end{table}

The exploratory design above, guided by the $F1_{cv}$ signal across methods, base models and folds, produced the configurations in Table~\ref{tab:progression}. The rest of this section asks three questions of the winning 9V---where do its per-class errors concentrate, what does the third specialist arbitrate and how much of the gain depends on augmentation.

\subsection{Per-Class Analysis}
\label{sec:per_class}

\begin{figure}[t]
\centering
\includegraphics[width=\columnwidth]{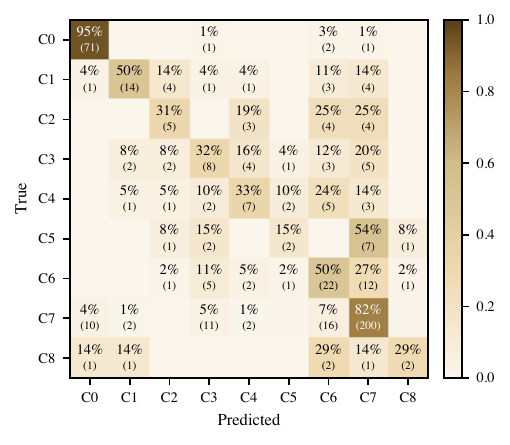}
\caption{Row-normalised confusion matrix of the winning 9V system on the hidden test set. C7~High-Adaptive absorbs most misclassifications from mid-hierarchy classes.}
\label{fig:confusion}
\end{figure}

On the hidden test set ($n{=}472$), the winning 9V performs strongly on surface-identifiable defences (C0~No~Defence $F1_{test}{=}.899$, C7~High-Adaptive $.833$) but struggles where categories overlap semantically (Figure~\ref{fig:confusion}, Appendix Table~\ref{tab:perclass}). The C0-override (Equation~\ref{eq:voting}) fires on $17.6\%$ of test samples, close to the $15.9\%$ training prevalence.

Two error patterns dominate. C6 and C7 are swapped on 28 samples (16 C6$\to$C7, 12 C7$\to$C6) and 7 of 13 C5~Neurotic samples ($54\%$) are labelled C7~High-Adaptive---the highest relative confusion rate. All three classes produce measured, reflective language and distinguishing them needs intent or longitudinal context rather than a single utterance \citep{perry2014anomalies}. The model therefore defaults to C7---the clinically costly direction, where a neurotic defence read as mature coping misses the signal for intervention.

\subsection{Voter Diversity: Flipping and Arbitration}
\label{sec:voter_diversity}

\begin{figure}[h!]
\centering
\includegraphics[width=\columnwidth]{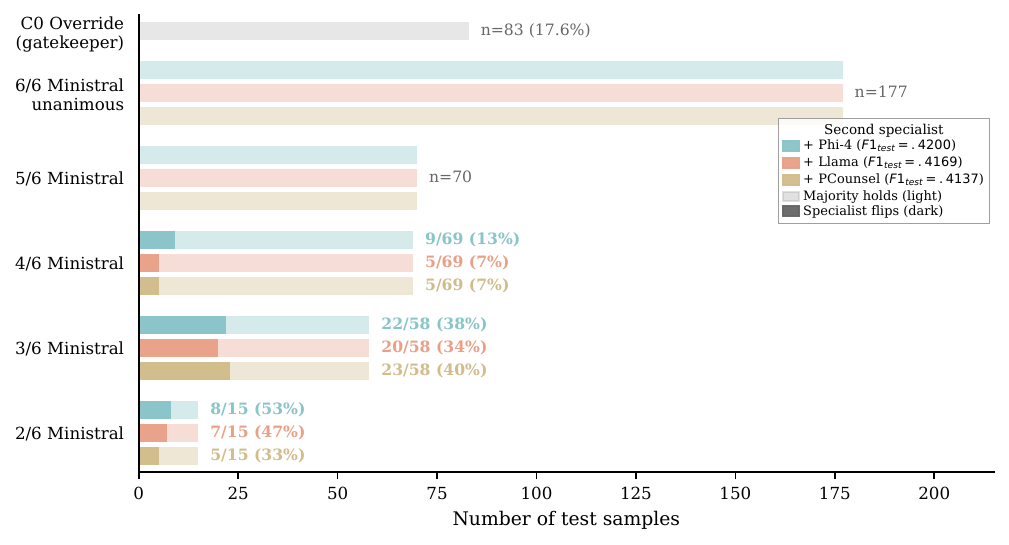}
\caption{Each bar is one 9V system (6 Ministral + 3 specialist voters); y-axis groups samples by how many Ministral voters agreed---the specialist can flip the Ministral majority only from 4/6 downwards---and dark portions mark the actual flips.}
\label{fig:voting_flow}
\end{figure}

In the 9V the third specialist cannot overrule a confident Ministral majority since 5/6 and 6/6 are mathematically locked against 3 specialist votes, so it can only intervene on the $n{=}142$ samples where Ministral is itself split (Figure~\ref{fig:voting_flow}). The question is whether the specialist flips Ministral's wrong calls (helpful arbitration) or its correct ones (harmful noise). Adding Phi4-LR\,8c lowers the system Krippendorff's $\alpha$ from $.451$ (6V Min-SFT\,9c + Min-LR\,8c) to $.397$ (9V) and the drop sits across branches rather than within them (within-branch Min-SFT $.617$, Min-LR $.630$, Phi4-LR $.464$; lowest cross-pair Min-SFT$\times$Phi4-LR $.382$). Such cross-branch disagreement among accurate base models is the well-known prerequisite for ensemble gains beyond the strongest member \citep{dietterich2000ensemble}. But low Krippendorff's $\alpha$ only proves the voters disagree, not whether they disagree where it matters, so we trace the actual flips.

Phi4-LR flips Ministral on 39 of those 142 samples and 33 ($85\%$) touch the C6/C7 boundary that dominates the per-class confusion matrix (Figure~\ref{fig:confusion}). C7 is the most common flip source (12 of 39, $31\%$), reflecting Min's tendency to over-call the majority class, but the redistribution is heterogeneous with no single direction dominating (top flip C7$\to$C3 is 6 of 39). Swapping Phi4-LR for Llama-LR or PCounsel-LR yields $.417$ and $.414$, so the gain reproduces across all three candidates.

\subsection{Augmentation Ablation}
\label{sec:aug_ablation}

Without GPT-5.2 augmentation (\emph{no-aug}; Table~\ref{tab:aug_ablation}), the 9V loses only $.042$.
Augmentation alone hurts a single model ($-.008$), but voting averages over the synthetic noise and turns augmentation into a $+.042$ lift on top of voting's $+.063$---they are interlocked, not additive.

\begin{table}[h!]
  \centering
  \footnotesize
  \setlength{\tabcolsep}{4pt}
  \begin{tabular*}{\columnwidth}{@{\extracolsep{\fill}}lcc@{}}
    \toprule
    \textbf{System} & \textbf{aug} & \textbf{no-aug} \\
    \midrule
    Min-SFT\,9c (single, no voting) & $.307$ & $.315$ \\
    5V Min-SFT\,9c                  & $.373$ & $.319$ \\
    + Min-LR\,8c (6V)               & $.391$ & $.369$ \\
    + Phi4-LR\,8c (9V)              & $\boldsymbol{.420}$ & $.378$ \\
    \bottomrule
  \end{tabular*}
  \caption{Augmentation ablation ($F1_{test}$, classes 1--8); augmentation hurts a single model but helps once voting averages the noise.}
  \label{tab:aug_ablation}
\end{table}

\section{Conclusion}
\label{sec:conclusion}

Our 9-voter ensemble reaches $F1_{test}{=}.420$ ($+33.4\%$ over baseline), driven by voter error independence. C0 forms a well-separated cluster while the defence-class clusters are less distinct---exactly the fuzzy boundaries our approach disentangles. We got there by stacking diversity axes (class granularity, training method, base model) and following the $F1_{cv}$ signal. GPT-5.2 augmentation hurts a single model alone but lifts the 9V by $+.042$. Post-hoc analysis reveals aug~$\times$~no-aug as another diversity axis---a 6V combining aug Min-SFT\,9c and no-aug PCounsel-LR\,8c clears $.40$ (Appendix Table~\ref{tab:posthoc}), supporting our hypothesis that independent errors are the lever. Voter diversity tends to help but does not scale arbitrarily. Disentangling these clusters with richer signal and matching the 9V more cheaply remain open.

\newpage
\section*{Limitations}

PSYDEFCONV provides only 1{,}864 training samples, so the $+.029$ gain from the best non-cross-model 6V ($.391$) to the 9V ($.420$) is a single hidden-test observation and what generalises is the complementary-model selection principle, not the exact 9V ranking. Several design choices---specialist selection by an $n{=}5$ Pearson correlation, top-3 fold selection without cross-validation and the C0-override threshold---rest on limited statistical support, so the submitted configuration is one of several plausible winners. Moderate annotator agreement ($\kappa{=}.639$) bounds the macro-F1 target and places the rare clinical classes C2, C5 and C8 inside the annotator-disagreement band. All evaluation is on PSYDEFCONV/ESConv (English) with 738 GPT-5.2 synthetic dialogues that carry generator-specific artefacts. The 9V needs distillation for real-time deployment, though the 5V single-branch already captures more than half of the gain at one third of the inference cost.

\section*{Ethics Statement}

At $F1_{test}{=}.420$ the system misclassifies the majority of defence-bearing utterances with an adaptive-skew bias that under-flags exactly the patients who most warrant clinical attention. Assigning defence labels to a person's utterances is itself a psychological intervention and should not occur outside supervised clinical workflows with informed consent \citep{steigerwald2026ethics,na-etal-2025-survey}. Like other LLM-based tools in mental health, such systems should augment, not replace, the human practitioner \citep{steigerwald2025caia}; outputs are categorical labels only and could be misused to pathologise individuals in adversarial contexts, mirroring known privacy, bias and accountability risks of mental-health LLMs \citep{steigerwald2026fromhelp}. The DMRS taxonomy reflects Western, English-language therapeutic traditions and PSYDEFCONV \citep{liu2021esconv} is simulated rather than clinical data whose crowdworkers consented to support-dialogue collection, not to psychodynamic re-annotation.

\section*{Data Availability}

The 738 GPT-5.2 synthetic dialogues, generation prompt and parameters are released under CC BY-NC 4.0 at \url{https://github.com/th-nuernberg/nuernberg-nlp-psydefdetect}.

\bibliography{references}

\appendix

\section{Hyperparameter Details}
\label{sec:hyperparams}

All fine-tuning uses 4-bit NF4 QLoRA on all linear projections (dropout $0.05$, cosine schedule with $10\%$ warm-up, 10 epochs, effective batch size 8, max sequence length $4{,}096$ tokens). SFT uses LoRA rank 32 ($\alpha{=}64$), learning rate $10^{-4}$, cross-entropy on the label digit. ClsHead uses rank 16 ($\alpha{=}32$), learning rate $2{\times}10^{-5}$, focal loss \citep{lin2017focal} ($\gamma{=}2$) with inverse-frequency class weights $w_c = N/(K\,n_c)$. LR is L2-regularised multinomial logistic regression on frozen last-token hidden states ($C \in \{10^{-3},10^{-2},10^{-1},1,10\}$ swept per fold; \texttt{class\_weight="balanced"}; \texttt{scikit-learn} defaults). Augmented samples enter only training.

\begin{tcolorbox}[
  colback=clrEns!10!white,
  colframe=clrEns!80!black,
  title={\fontsize{8}{9.5}\selectfont\bfseries SFT Prompt Template},
  fontupper=\fontsize{7.5}{9}\selectfont,
  breakable,
  boxrule=0.5pt,
  left=4pt, right=4pt, top=3pt, bottom=3pt,
  toptitle=2pt, bottomtitle=2pt,
]
\textbf{System:} You are an expert psychologist specialising in the Defense Mechanism Rating Scale (DMRS). You analyse emotional support conversations and classify the psychological defence mechanisms used by the help-seeker in their utterances. You always respond with exactly one line in the format `label: \textless{}number\textgreater' where \textless{}number\textgreater{} is 1--8.\\[4pt]
\textbf{User:} Below is an emotional support conversation between a SEEKER and a SUPPORTER. Your task is to classify the TARGET utterance according to the Defense Mechanism Rating Scale (DMRS).\\[4pt]
\textbf{\#\# Conversation}\\
{\fontsize{7}{8.5}\selectfont\textit{[Full dialogue history with SEEKER and SUPPORTER turns]}}\\[2pt]
\textbf{\#\# Target Utterance}\\
{\fontsize{7}{8.5}\selectfont\textit{[The seeker utterance to be classified]}}\\[2pt]
\textbf{\#\# DMRS Defence Mechanism Categories}\\
{\fontsize{7}{8.5}\selectfont\textit{[Definitions for all 8 (or 9) DMRS levels, e.g.\ ``1: Action: The speaker uses action-oriented defences such as acting out, passive aggression, or help-rejecting complaining\ldots''; full descriptions for every level]}}\\[4pt]
Examine the dialogue carefully and select the single most appropriate defence tier. When multiple defences seem plausible, choose the tier with the strongest supporting evidence. Every utterance contains a defence mechanism. Return exactly one line: \texttt{label: \textless{}1--8\textgreater{}}.\\[4pt]
\textbf{Assistant:} \texttt{label: \{class\_id\}}
\end{tcolorbox}

\section{Multi-Model CV5 Comparison}
\label{sec:cv5_app}

Table~\ref{tab:cv5_models} reports $F1_{cv}$ for all candidate base models, methods and class modes.
The LR column dominates across models and Ministral-8B and Phi-4-14B top the $8$-class LR column---the pairing adopted in the winning system.

\begin{table}[ht]
  \centering
  \footnotesize
  \setlength{\tabcolsep}{2.8pt}
  \begin{tabular*}{\columnwidth}{@{\extracolsep{\fill}}lcccccc@{}}
    \toprule
    & \multicolumn{2}{c}{\textbf{SFT}} & \multicolumn{2}{c}{\textbf{ClsHead}} & \multicolumn{2}{c}{\textbf{LR}} \\
    \cmidrule(lr){2-3}\cmidrule(lr){4-5}\cmidrule(lr){6-7}
    \textbf{Model} & 8c & 9c & 8c & 9c & 8c & 9c \\
    \midrule
    \textbf{Ministral-8B}    & .321 & .306 & .333 & .311 & \textbf{.342} & .315 \\
    \textbf{Phi-4-14B}       &  --  & .293 & .337 &  --  & \textbf{.337} &  --  \\
    Llama-3.1-8B             & .251 & .279 & .246 & .284 & .312 & .284 \\
    Qwen2.5-7B               & .266 & .256 & .302 & .268 & .307 & .283 \\
    PsychoCounsel-8B         &  --  &  --  & .316 &  --  & .301 &  --  \\
    PsyLLM-8B                &  --  &  --  & .295 &  --  & .289 &  --  \\
    GPT-OSS-20B              & .212 & .183 & .278 &  --  & .292 &  --  \\
    \bottomrule
  \end{tabular*}
  \caption{$F1_{cv}$ for all candidate base models (mean over $n{=}5$ folds, classes 1--8; $\sigma{\in}[.019,.045]$).}
  \label{tab:cv5_models}
\end{table}

\section{Data Augmentation and Class Balancing}
\label{sec:augmentation}

GPT-5.2 (temperature $0.9$) generated 738 synthetic dialogues against the 80/20 dialog-stratified train split (1{,}520 originals) using a few-shot prompt with the full DMRS taxonomy and five randomly sampled originals of the target class; each dialogue is a 2--6 turn emotional support exchange ending in a target seeker utterance that demonstrates the specified defence level.
Table~\ref{tab:augmentation} reports the per-class budget against this 80/20 split. The same 738 synthetic dialogues underpin both the ``Min-SFT\,9c full-train, augmented (single model)'' baseline in Table~\ref{tab:progression} (trained on all 1{,}864 originals plus the 738 synthetic) and every CV5 voter (each trained on its fold-train split of $\sim$1{,}493 originals plus the same 738 synthetic).
Synthetic data enters only training splits, while validation and test sets remain exclusively original human-annotated data.

\begin{tcolorbox}[
  colback=clrEns!10!white,
  colframe=clrEns!80!black,
  title={\fontsize{8}{9.5}\selectfont\bfseries Data Augmentation Prompt (Abridged)},
  fontupper=\fontsize{7.5}{9}\selectfont,
  breakable,
  boxrule=0.5pt,
  left=4pt, right=4pt, top=3pt, bottom=3pt,
  toptitle=2pt, bottomtitle=2pt,
]
\textbf{System:} You are a psychology expert generating training data. Output only valid JSON.\\[4pt]
\textbf{User:} You are an expert psychologist specialising in psychological defence mechanisms.\\[2pt]
{\fontsize{7}{8.5}\selectfont\textit{[Full DMRS taxonomy with definitions and example markers for all 9 levels]}}\\[4pt]
\textbf{Your Task:} Generate \{$n$\} NEW and DIVERSE examples of \textbf{Level \{$\ell$\}: \{name\}} in emotional support conversations.\\[4pt]
\textbf{Requirements:}
\begin{enumerate}[nosep,leftmargin=12pt,label=\arabic*.]
  \item Create REALISTIC dialogues between a help-seeker and emotional supporter
  \item The TARGET UTTERANCE must clearly demonstrate Level \{$\ell$\}
  \item VARY the topics: work stress, relationships, health anxiety, family conflict, finances, grief, \ldots
  \item Use natural, conversational English
  \item The SEEKER uses the defence mechanism (not the supporter)
  \item Dialogue context: 2--6 turns before the target
\end{enumerate}
\vspace{2pt}
\textbf{Few-Shot Examples:}\\
{\fontsize{7}{8.5}\selectfont\textit{[5 randomly sampled original training examples of the target class, each showing dialogue context + target utterance + label]}}\\[4pt]
\textbf{Output Format:} Generate exactly \{$n$\} examples as JSON objects, each with a dialogue (list of speaker/text turns) and a target utterance.
\end{tcolorbox}

\begin{table}[ht]
  \centering
  \footnotesize
  \setlength{\tabcolsep}{4pt}
  \begin{tabular*}{\columnwidth}{@{\extracolsep{\fill}}clrrr@{}}
    \toprule
    \textbf{ID} & \textbf{Defence Level} & \textbf{Orig.} & \textbf{+Aug} & \textbf{Total} \\
    \midrule
    0 & No Defence         & 244 &   0 & 244 \\
    1 & Action             &  88 & 112 & 200 \\
    2 & Major Image-Dist.\ &  54 & 146 & 200 \\
    3 & Disavowal          &  83 & 117 & 200 \\
    4 & Minor Image-Dist.\ &  67 & 133 & 200 \\
    5 & Neurotic           &  34 & 102 & 136 \\
    6 & Obsessional        & 135 &  65 & 200 \\
    7 & High-Adaptive      & 794 &   0 & 794 \\
    8 & Needs More Info    &  21 &  63 &  84 \\
    \midrule
      & \textit{Total}     & \textit{1{,}520} & \textit{738} & \textit{2{,}258} \\
    \bottomrule
  \end{tabular*}
  \caption{Per-class composition of the 80/20 train split against which the augmentation budget was computed; C0 and C7 are excluded from augmentation but remain in training.}
  \label{tab:augmentation}
\end{table}

\section{Embedding Geometry after 8-Class Specialist Training}
\label{sec:emb_app}

\begin{figure}[t]
\centering
\includegraphics[width=\columnwidth]{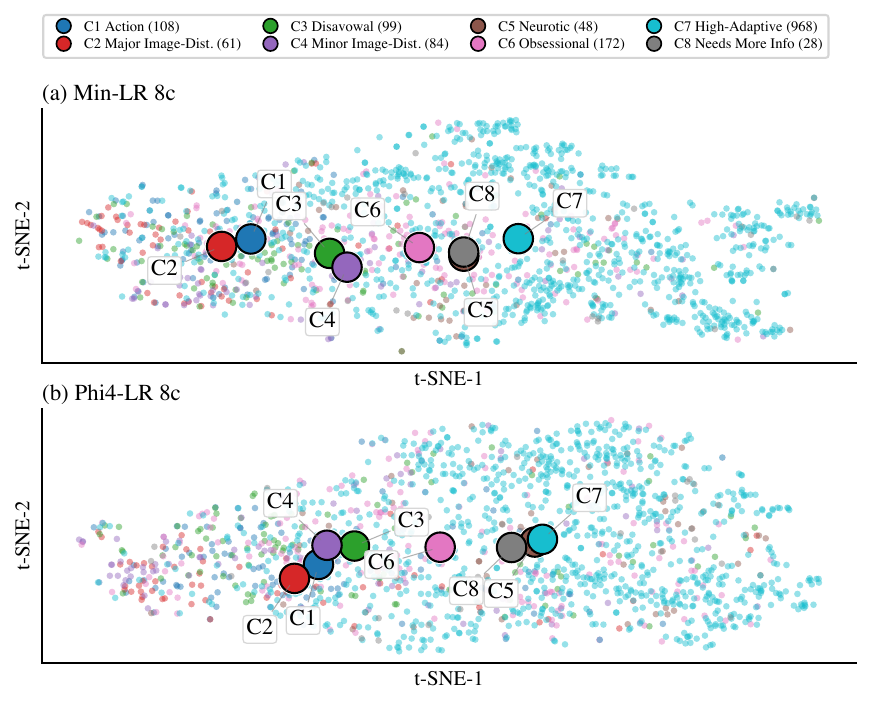}
\caption{Per-class t-SNE of the LR\,8c specialist hidden states ($n{=}1{,}568$, C0 excluded).}
\label{fig:emb_8c}
\end{figure}

Figure~\ref{fig:emb_8c} shows per-class t-SNEs for the two LR\,8c specialists. 8-class training does not separate the defence clusters in either model---C6 and C7 remain the dominant overlap (cf.\ Figure~\ref{fig:confusion})---yet the two models produce visibly different local geometries, consistent with our finding that error independence drives the ensemble.

\section{Per-Class Results}
\label{sec:perclass_app}

Table~\ref{tab:perclass} reports per-class scores of the 9V ensemble, with C3 and C5 below $.300$ and confirming the C7 absorption seen in Figure~\ref{fig:confusion}.

\begin{table}[t]
  \centering
  \scriptsize
  \setlength{\tabcolsep}{3pt}
  \begin{tabular*}{\columnwidth}{@{\extracolsep{\fill}}clcccc@{}}
    \toprule
    \textbf{ID} & \textbf{Defence Level} & $\boldsymbol{F1}$ & \textbf{P} & \textbf{R} & $\boldsymbol{n}$ \\
    \midrule
    0 & No Defence         & .899 & .855 & .947 &  75 \\
    1 & Action             & .583 & .700 & .500 &  28 \\
    2 & Major Image-Dist.  & .333 & .357 & .312 &  16 \\
    3 & Disavowal          & .291 & .267 & .320 &  25 \\
    4 & Minor Image-Dist.  & .350 & .368 & .333 &  21 \\
    5 & Neurotic           & .200 & .286 & .154 &  13 \\
    6 & Obsessional        & .436 & .386 & .500 &  44 \\
    7 & High-Adaptive      & .833 & .844 & .823 & 243 \\
    8 & Needs More Info    & .333 & .400 & .286 &   7 \\
    \bottomrule
  \end{tabular*}
  \caption{Per-class $F1_{test}$ of the winning 9V ensemble on the hidden test set ($n{=}472$).}
  \label{tab:perclass}
\end{table}

\section{Post-Hoc Re-Voting Search}
\label{sec:posthoc}

The configurations in this appendix are not part of our shared-task submission and were never uploaded to the leaderboard.
After the test labels were released we re-voted over our 18 cached per-fold prediction sets (12 aug + 6 no-aug) without retraining; an 18-branch search takes seconds on cached predictions.
Fold selection is post-hoc, so these scores are oracle upper bounds rather than blind submissions.

Three patterns emerge (Table~\ref{tab:posthoc}). First, mixing aug with no-aug branches is itself a diversity axis: 9V ensembles combining aug and no-aug specialists average $F1_{test}{=}.391$ vs.\ $.372$ for aug-only pairs and every top-3 entry from 9V upward draws on at least one no-aug branch. Second, for 6V the best gatekeeper~+~specialist pairing is the no-aug PCounsel-LR\,8c ($.402$), narrowly ahead of no-aug Phi4-LR\,8c ($.396$); the cross-architecture LR specialists dominate the top of the 6V leaderboard. Third, ensemble gain grows from 9V to 12V (best $.452 \to .471$, $+.019$) and begins to plateau beyond 12V, suggesting that adding more branches from a saturated voter pool re-introduces correlated errors faster than independent signal.

\begin{table}[t]
  \centering
  \scriptsize
  \setlength{\tabcolsep}{2.5pt}
  \begin{tabular*}{\columnwidth}{@{\extracolsep{\fill}}ccl@{}}
    \toprule
    $\boldsymbol{F1_{test}}$ & $\boldsymbol{t}$ & \textbf{Configuration (gatekeeper + specialists)} \\
    \midrule
    \multicolumn{3}{@{}l}{\textit{6V}} \\
    $.402$ & 2 & Min-SFT\,9c + PCounsel-LR\,8c (n) \\
    $.396$ & 2 & Min-SFT\,9c + Phi4-LR\,8c (n) \\
    $.395$ & 2 & Min-SFT\,9c + Min-LR\,8c \\
    \midrule
    \multicolumn{3}{@{}l}{\textit{9V}} \\
    $.452$ & 1 & Min-SFT\,9c + Min-SFTinit-LR\,8c + Phi4-LR\,8c (n) \\
    $.449$ & 2 & Min-SFT\,9c + Min-SFTinit-LR\,8c + Phi4-LR\,8c (n) \\
    $.445$ & 3 & Min-SFT\,9c + Min-SFTinit-LR\,8c + Phi4-LR\,8c (n) \\
    \midrule
    \multicolumn{3}{@{}l}{\textit{12V}} \\
    $.471$ & 3 & Phi4-SFT\,9c + Min-Cls\,9c + Min-LR\,8c (n) + Phi4-LR\,8c (n) \\
    $.464$ & 3 & Phi4-SFT\,9c + Llama-LR\,8c + Min-LR\,8c (n) + Phi4-LR\,8c (n) \\
    $.456$ & 3 & Phi4-SFT\,9c + Min-SFTinit-LR\,8c + Min-LR\,8c (n) + Phi4-LR\,8c (n) \\
    \bottomrule
  \end{tabular*}
  \caption{Top-3 post-hoc voter combinations on the test set per ensemble size; (n) marks no-aug branches (unmarked = augmented), $t$ is the C0-override threshold; Min-SFTinit-LR is a Min-LR\,8c specialist initialised from the SFT\,9c adapter rather than ClsHead.}
  \label{tab:posthoc}
\end{table}

\end{document}